\documentclass[sigconf]{acmart}
\usepackage{enumitem}
\usepackage{graphicx}
\usepackage{adjustbox}
\usepackage{multirow}
\usepackage{balance} 

\usepackage{soul}

\AtBeginDocument{%
  }

\copyrightyear{2024}
\acmYear{2024}
\setcopyright{acmlicensed}\acmConference[CIKM '24]{Proceedings of the 33rd ACM International Conference on Information and Knowledge Management}{October 21--25, 2024}{Boise, ID, USA}
\acmBooktitle{Proceedings of the 33rd ACM International Conference on Information and Knowledge Management (CIKM '24), October 21--25, 2024, Boise, ID, USA}
\acmDOI{10.1145/3627673.3679218}
\acmISBN{979-8-4007-0436-9/24/10}

\begin{document}

\title{Introducing CausalBench: A Flexible Benchmark Framework for Causal Analysis and Machine Learning}\titlenote{This work is supported by NSF grant 2311716, "CausalBench: A Cyberinfrastructure for Causal-Learning Benchmarking for Efficacy, Reproducibility, and Scientific Collaboration"}

\author{Ahmet Kapki\c{c}}
\email{akapkic@asu.edu}
\orcid{0000-0002-7142-093X}
\affiliation{%
  \institution{Arizona State University}
 \city{Tempe}
 \state{AZ}
 \country{USA}
}
\author{Pratanu Mandal}
\email{pmandal5@asu.edu}
\orcid{0009-0007-0529-2179}
\affiliation{%
  \institution{Arizona State University}
 \city{Tempe}
 \state{AZ}
 \country{USA}
}
\author{Shu Wan}
\email{swan@asu.edu}
\orcid{0000-0003-0725-3644}
\affiliation{%
  \institution{Arizona State University}
 \city{Tempe}
 \state{AZ}
 \country{USA}
}
\author{Paras Sheth}
\email{psheth5@asu.edu}
\orcid{0000-0002-6186-6946}
\affiliation{%
  \institution{Arizona State University}
 \city{Tempe}
 \state{AZ}
 \country{USA}
}
\author{Abhinav Gorantla}
\email{agorant2@asu.edu}
\orcid{0009-0003-1242-5671}
\affiliation{%
  \institution{Arizona State University}
 \city{Tempe}
 \state{AZ}
 \country{USA}
}  

\author{Yoonhyuk Choi}
\email{ychoi139@asu.edu}
\orcid{0000-0003-4359-5596}
\affiliation{%
  \institution{Arizona State University}
 \city{Tempe}
 \state{AZ}
 \country{USA}
}

\author{Huan Liu}
\email{huanliu@asu.edu}
\orcid{0000-0002-3264-7904}
\affiliation{%
  \institution{Arizona State University}
 \city{Tempe}
 \state{AZ}
 \country{USA}
}

\author{K. Sel\c{c}uk Candan}
\email{candan@asu.edu}
\orcid{0000-0003-4977-6646}
\affiliation{%
  \institution{Arizona State University}
 \city{Tempe}
 \state{AZ}
 \country{USA}
}



\renewcommand{\shortauthors}{Ahmet Kapkiç et al.}


\begin{abstract}

While witnessing the exceptional success of machine learning (ML) technologies in many applications, users are starting to notice a critical shortcoming of ML: correlation is a poor substitute for causation. The conventional way to discover causal relationships is to use randomized controlled experiments (RCT); in many situations, however, these are impractical or sometimes unethical. Causal learning from observational data offers a promising alternative. While being relatively recent, causal learning aims to go far beyond conventional machine learning, yet several major challenges remain. Unfortunately,  advances are hampered due to the lack of unified benchmark datasets, algorithms, metrics, and evaluation service interfaces for causal learning. In this paper, we introduce {\em CausalBench}, a transparent, fair, and easy-to-use evaluation platform, aiming to (a) enable the advancement of research in causal learning by facilitating scientific collaboration in novel algorithms, datasets, and metrics and (b) promote scientific objectivity, reproducibility, fairness, and awareness of bias in causal learning research. CausalBench provides services for benchmarking data, algorithms, models, and metrics, impacting the needs of a broad of scientific and engineering disciplines. 
\end{abstract}



\maketitle

\section{Introduction \label{sec:introd} }
\begin{figure}[t]
    \centering
    \includegraphics[width=0.9\columnwidth]{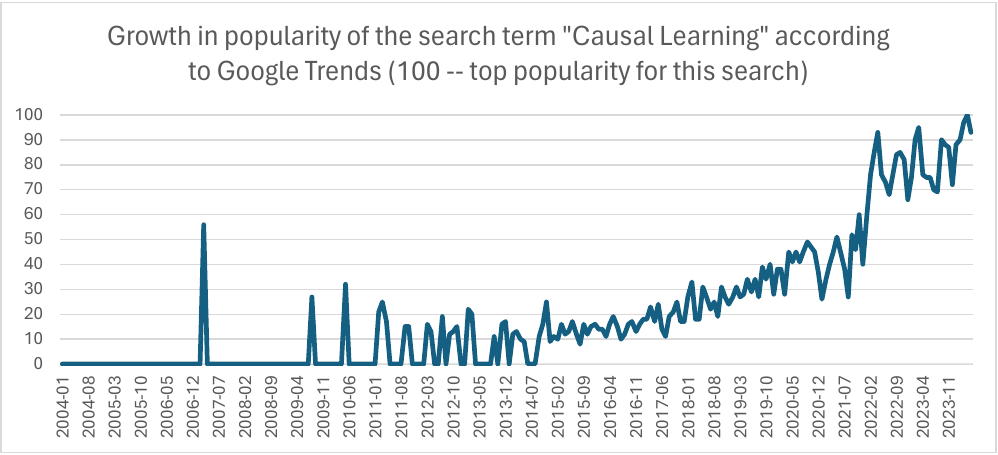}
    \caption{Causal learning has exploded in popularity in recent years}
    \label{fig:cl_op}
\end{figure}

Machine learning (ML) is serving as a key pillar in scientific innovation~\cite{cockburn2018impact} in a myriad of high-impact science domains, such as medical science~\cite{MlMed}, epidemiology~\cite{MlEpi}, and environmental health~\cite{EnvSus}. Nevertheless, users are starting to notice a critical shortcoming of the traditional ML techniques, which can learn correlation-based patterns from data (Figure \ref{fig:cl_op}): data may contain spurious correlations and correlation is a poor substitute for causation \cite{Simon1977}.

Consequently, successfully tackling many urgent challenges in socio-economically critical domains requires a deeper understanding of causal relationships and interactions from observational data, and causal learning offers a promising alternative to correlation-based learning \cite{hofler2005,Prosperi2020, azad2024vision}. For example, developing a plan for combining, co-operating, and designing portfolios of natural and built water infrastructure requires an understanding of the causally complex interplay of entities in a multi-layer network, including physics underlying natural as well as built infrastructures for flood protection, erosion control, water storage, and purification\footnote{This research has been funded by a US Army Corps of Engineers Engineering With Nature Initiative through Cooperative Ecosystem Studies Unit Agreement $\#$W912HZ-21-2-0040}~\cite{causalhydro, Sheth2023STREAMS, Sheth2022STCD}.

Standardized evaluation played a major role in ML development and contributed to the impressive impact of ML in scientific innovation. Successful early benchmarking efforts, such as UCI ML and UCI KDD repositories \cite{UCIML, UCIKDD, Sheth2021CauseBox}, not only helped guide the development of efficient and effective ML algorithms but also encouraged
collaborative research and paved the way for recent breakthroughs in deep learning. For example, to evaluate an image classifier, we have widely used metrics (e.g., accuracy, 
F1 score,  ROC-AUC~\cite{ROCAUC}), procedures (e.g., cross-validation~\cite{CrossValid}), and datasets (e.g., MNIST~\cite{deng2012mnist}, CIFAR10~\cite{CIFAR10}, ImageNet~\cite{ImageNet}). More recent frameworks, such as~\cite{OpenML}, move towards a collaborative approach, where datasets, models, and metrics are provided by the members of the community.  

\begin{table}[t]
    \caption{Popular causal ML tools with the supported data, methods, and metrics~\cite{MLTable}}
    \label{tab:sota}
    \centering
\begin{adjustbox}{width=0.9\columnwidth,height=0.3\columnwidth}
    \begin{tabular}{|cc|ccc|cc|cc|}
    \hline
          &       & \multicolumn{3}{c|}{\textbf{C. Effect Estim.}} & \multicolumn{2}{c|}{\textbf{C. Discovery}} & \multicolumn{2}{c|}{\textbf{Evaluation}} \\
          &       & \textbf{CausalML} & \textbf{EconML} & \textbf{DoWhy} & \textbf{CausalNex} & \textbf{CausalDisc.} & \textbf{C-Benchm.} & \textbf{JustCause} \\
    \hline
    \multirow{4}{*}{\rotatebox{90}{\textbf{Data}}} & \textbf{iid} & x     & x     & x     & x     & x     & x     & x \\
          & \textbf{IV} & x     & x     & x     &       &       &       &  \\
          & \textbf{Graph} &       &       &       &       &       &       &  \\
          & \textbf{T.Series} &       &       & x     &       &       &       &  \\
    \hline
    \multirow{9}{*}{\rotatebox{90}{\textbf{Methods}}} & \textbf{Pro. Score} &       & x     & x     &       &       &       & x \\
          & \textbf{Tree-based} & x     & x     & x     &       &       &       &  \\
          & \textbf{Meta-Learn.} & x     & x     & x     &       &       &       & x \\
          & \textbf{Doubly ML} &       & x     & x     &       &       &       &  \\
          & \textbf{D.Robust} &       & x     & x     &       &       &       & x \\
          & \textbf{IV} & x     & x     & x     &       &       &       &  \\
          & \textbf{Mediation} &       &       & x     &       &       &       &  \\
          & \textbf{Graph} &       &       &       & x     & x     &       &  \\
          & \textbf{Pairwise} &       &       &       &       & x     &       &  \\
    \hline
    \multirow{11}{*}{\rotatebox{90}{\textbf{Metrics}}} & \textbf{PEHE} &       & \multirow{11}[2]{*}{\rotatebox{90}{User Provided Metrics}} &       &       &       &       & x \\
          & \textbf{RMSE} & x     &       &       &       &       & x     & x \\
          & \textbf{MAE} & x     &       &       &       &       &       & x \\
          & \textbf{BIAS} &       &       &       &       &       & x     & x \\
          & \textbf{Coverage} &       &       &       &       &       & x     &  \\
          & \textbf{Conf.Int.} &       &       &       &       &       & x     &  \\
          & \textbf{Agg.Score} &       &       &       &       &       & x     &  \\
          & \textbf{Refut.} &       &       & x     &       &       &       &  \\
          & \textbf{SID} &       &       &       &       & x     &       &  \\
          & \textbf{SHD} &       &       &       &       & x     &       &  \\
          & \textbf{Classif.} &       &       &       & x     & x     &       &  \\
    \hline
    \end{tabular}
\end{adjustbox}
\end{table}

In this paper, we argue that the causal
learning community can achieve the same by meticulously surveying the emerging field
of vibrant research, systematically categorizing the existing benchmarking efforts into technically meaningful groups, and discovering the areas where further efforts are in dire need. While initial work in this area has started (Table~\ref{tab:sota}), more systematic advances are required. Shared datasets and metrics for benchmarking can be extremely valuable for not only causal learning algorithm design, but also for comparison and benchmarking of available solutions. Currently, only a fraction of existing studies are replicable and with each version of a GPU driver or a Python library, performance results can vary wildly. Yet, despite the promise of advancing science and research, such data can be difficult to find and costly to annotate. Here, we argue that the recent availability of big observational data in all walks of life offers us an unprecedented opportunity to consolidate the hitherto distributed and unorganized efforts by creating a cyberinfrastructure for advancing causal learning research.
 
Based on this premise, here we introduce the {\em CausalBench} platform, a novel cyberinfrastructure for benchmarking causal learning. Aiming for a systematic, objective, and transparent evaluation of causal learning models and algorithms, CausalBench integrates publicly available benchmarks and consensus-building standards for the evaluation of causal learning models and algorithms from observational data. Consisting of a publicly accessible data and algorithm repository along with service APIs, the platform assists researchers and developers in easily applying and effectively evaluating (a) causal inference, (b) causal discovery, and (c) causal interpretability algorithms with a variety of standard metrics, procedures, and large-scale datasets.

In the rest of this paper, we first discuss the principles that are the pillars of CausalBench (Section \ref{sec:princ}). We then provide an overview of the framework and its functionalities (Section \ref{sec:framework}). In Section \ref{sec:demo}, we discuss usage scenarios of the system.
\section{Key Objectives and Design Principles \label{sec:princ} } 
As a  platform to systematically and reliably benchmark causal learning models and algorithms, CausalBench aims to target the following key objectives: 

\begin{itemize}[leftmargin=*]
\item{\em Objective \#1: Universally adopted metrics, procedures, and datasets.} This involves conducting an extensive identification of existing datasets, performance metrics, and procedures used in the evaluation of state-of-the-art causal learning algorithms, and developing an “ontology” for benchmarking to standardize the evaluation methodology, improve transparency, and promote collaboration to advance causal learning efficiently.
\item {\em Objective \#2: A standard and convenient way for the community to contribute data and models.} Different from datasets for conventional machine learning, it is often difficult to obtain the ground truth of the causal relations among observed variables, not to mention the potential existence of unobserved variables -- in many cases, we have to work with datasets with incomplete causal knowledge. We need to make it easy and convenient for the community to contribute new data and models. 
\item {\em Objective \#3: Trustable (transparent, reproducible)  benchmarking.} In addition to making data, models, and metrics available to the researchers, the system should enable trustable, fair, reproducible, and open benchmarking of the available models and algorithms. In particular, all steps of an executed experiment, including the data, hyperparameters, as well as hardware/software configuration must be recorded and made transparently available to help support interpretation of the experiment results.
\item {\em Objective \#4: Fair and flexible comparisons of models and algorithms.}Conversely, one should be able to explore the results of recorded benchmark experiments and compare existing solutions fairly and flexibly. Fairness implies that if the models are compared, these models and the experiment settings must be compatible, and/or any differences in data, hyperparameters, and hardware/software settings that may impact the results must be highlighted. A fair system should account for biases caused by algorithms or system configurations. Flexibility means that the users of the system must be able to {\em slice-and-dice} the benchmark experiments in different ways, based on a different grouping or slicing criteria.
\end{itemize}

\begin{figure}[t]
    \centering
\includegraphics[width=0.75\columnwidth]{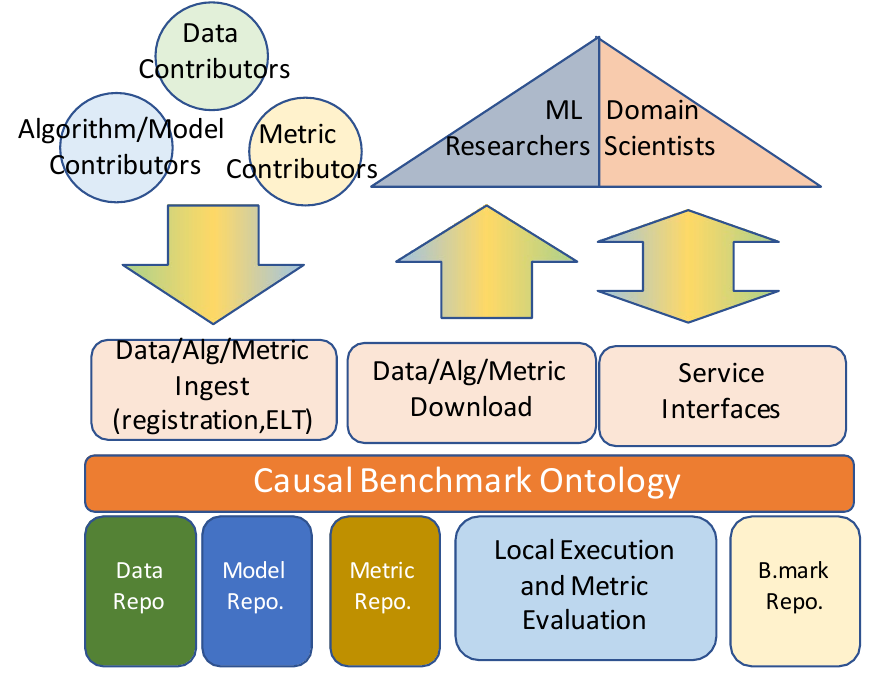}
    \caption{Overview of CausalBench}
    \label{fig:bmark}
    \vspace*{-0.15in}
\end{figure}

\section{CausalBench Framework \label{sec:framework}}

\subsection{Overall Architecture}
\begin{figure}[t]
   \centering
   \includegraphics[width=\columnwidth]{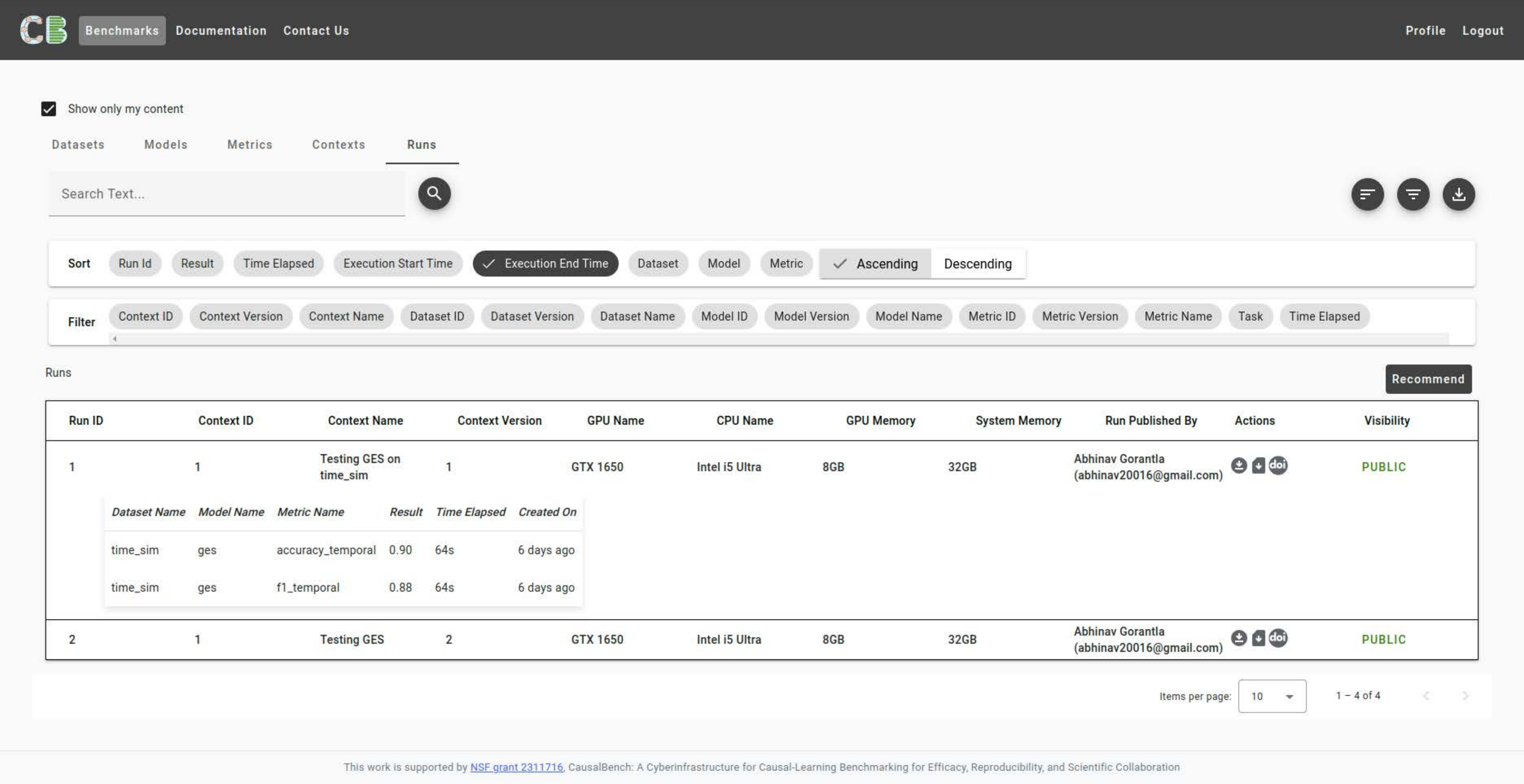}
   \caption{CausalBench runs page}
   \label{fig:appimg}
   \vspace{-0.1in}
\end{figure}

CausalBench is designed to enable its users to easily add new relevant datasets, models, and metrics (Figure~\ref{fig:bmark}). The platform boasts several key components:
\begin{itemize}[leftmargin=*]
\item A web-based dataset, model, and metric registration module provides a guided interface through which a provider registers a dataset, a model, or a metric with CausalBench. Registration involves the systematic acquisition of metadata needed for the discovery, access, and use of data and models. 

\item A data, model, and metric repository manages metadata associated with all registered datasets, models, and  metrics and ensures that these persist and are accessible. The repository further stores (a) benchmark contexts and experiment setups consisting of data, model, and metric components and (b) authenticated performance results of benchmark runs and the associated metadata (e.g., hyperparameters, hardware/software setups).
\item A benchmark runs page (Figure \ref{fig:appimg}) where performance results of runs, including results, system information, and a DOI attached to each benchmark run, is displayed. Experiment results are in a tabular format that can be sorted and filtered.
\item A CausalBench console-based Python package supports the execution of causal machine learning experiments. The package enables quantitative evaluation of the models (for accuracy and efficiency) based on datasets in the repository using local CPU and GPU resources.  
\item A web interface supports browsing through repositories of datasets, models, metrics and benchmark contexts, exploring (slice-and-dice) experiments across the runs executed through CausalBench. In addition to providing data download links and data descriptions, the platform also offers accessible APIs of evaluation metrics and service interfaces. 
\end{itemize}

\subsection{Benchmarking Causal ML Models}
CausalBench includes several core components. These include \textbf{datasets}, $\mathcal{D}$, which are  data files and configuration files that describe the properties of the data in the data files; \textbf{models}, $\mathcal{M}$, which are  algorithms written in Python that take in a dataset and execute a particular model, producing outputs based on the tasks and models; and \textbf{metrics} $\mathcal{A}$, which are Python implementation of metric calculations that take in the outputs provided by the model and output a numerical value, based on its configuration. CausalBench follows a flexible approach, where datasets, models, and metrics can be re-used for different causal machine learning tasks. The set of all causal machine learning tasks available at CausalBench is denoted as $\mathcal{T}$. Given the above, a \textbf{benchmark context}, $\mathcal{C}$, includes a subset (denoted by the subscript $\mathcal{}_P$) of datasets $\mathcal{D}$, models $\mathcal{M}$ and accuracy metrics $\mathcal{A}$, along with the appropriate parameter and hyperparameter settings:
\[
\mathcal{C} = \{(\mathcal{D}_P, \mathcal{M}_P, \mathcal{A}_P, \mathcal{H}_P), \mathcal{D}_P \subseteq \mathcal{D}, \; \mathcal{M}_P \subseteq \mathcal{M}, \; \mathcal{A}_P \subseteq \mathcal{A}\}.
\]
Above, $\mathcal{H}_P$ denotes the set of \textbf{parameter and hyperparameter settings} applicable to the execution or training of the models. Note that the benchmark context can equivalently be seen as a set of \textbf{benchmark scenarios}:
\[
\mathcal{C} = \{ (d, m, \mathcal{A}_P, h ) \mid d \in \mathcal{D}_P, \; m \in \mathcal{M}_P, \;
h \in \mathcal{H}_P  \}.
\]
An \textbf{instrumented context}, $\mathcal{I}$, is  a coupling of these benchmark scenarios with a particular user hardware/software system, $s$: 
\[
\mathcal{I}(\mathcal{C}, s) = \{ (d, m, \mathcal{A}_P, h , s ) \mid d \in \mathcal{D}_P, \; m \in \mathcal{M}_P, ; h \in \mathcal{H}_P  \}.
\]
A \textbf{benchmark run}, $\mathcal{R}(\mathcal{I}(\mathcal{C},s))$, then, is the recording of the outputs of the execution of the benchmark scenarios in an instrumented context, $\mathcal{I}$:
\[
\{
( A, T, S; d, m, h, s ) \mid (d, m, \mathcal{A}_P, h, s ) \in \mathcal{I}(\mathcal{C}, s) \},
\]
where $A$ is a set of key-value pairs recording the value for each accuracy metric $a \in \mathcal{A}_P$. $T$ is a set of key-value pairs recording the timing values for each timing metrics, such as {\em CPU-time}, {\em GPU-time}; and $S$ is a set of key-value pairs recording the system usage values for each resource metrics, such as {\em CPU-memory}, {\em GPU-memory}. Noting that the timing metrics $T$ and resource metrics $S$ are measured for each benchmark scenario. CausalBench stores authenticated benchmark runs of its users in public or private repositories and allows a user to compare multiple runs (that are accessible to them) of a task, dataset, and/or model.

\subsection{Reproducibility and Versioning}
In order to enable reproducible research on causal machine learning, once a dataset, model, or a metric is declared as public and is included in at least one public run, it becomes permanent in the system and cannot be removed. Benchmark runs that are made public are registered with an open-access repository, Zenodo, and are associated with a unique document object identifier (DOI). Of course, over time, datasets, models, and metrics may evolve. For any public, and therefore permanent, component, this involves the creation of a new version of the component, with its own unique identifier, maintained along with the old version.

\subsection{CausalBench Features}
The Python package for CausalBench is written in Python 3.10, and facilitates the creation, uploading, and executing of core components (a dataset, model, or metric) and contexts. Running CausalBench requires signing up to the system through the CausalBench website and providing the user credentials in the config file of the Python package. A user has several options available on launch: downloading/uploading a component, declaring and executing a benchmark run, and exploring existing benchmarks.

\paragraph{Exploring Data, Model, and Metric Repositories} Users can browse the repositories of available datasets, models, and metrics created by themselves or made public by other users. Each component is visualized as a card, providing an overview of the relevant statistics of the components.  Clicking on any card provides details and allows downloading the component. The cards corresponding to the versions of the same component are clustered and stacked.

\paragraph{Execution and Registration of Benchmark Runs} A benchmark run is essentially a benchmark scenario (a combination of datasets, models, and metrics) instrumented and executed on the user's local resources. The UI helps the user in the process of creating benchmark scenarios by filtering out incompatible components and highlighting suitable ones as the user starts declaring aspects of the benchmark scenario. This suggestion feature works based on the inputs and the outputs of each component and their task type. Executing a benchmark run includes creating an instance of the benchmark scenario with current system and environment configurations on the local machine, running configurations for each combination of the core components, and uploading the execution results, including the corresponding resource usage information back to CausalBench repositories. Once declared public, these results are registered as permanent and associated with DOIs.

\begin{figure}[t]
   \centering
   \includegraphics[width=0.8 \columnwidth]{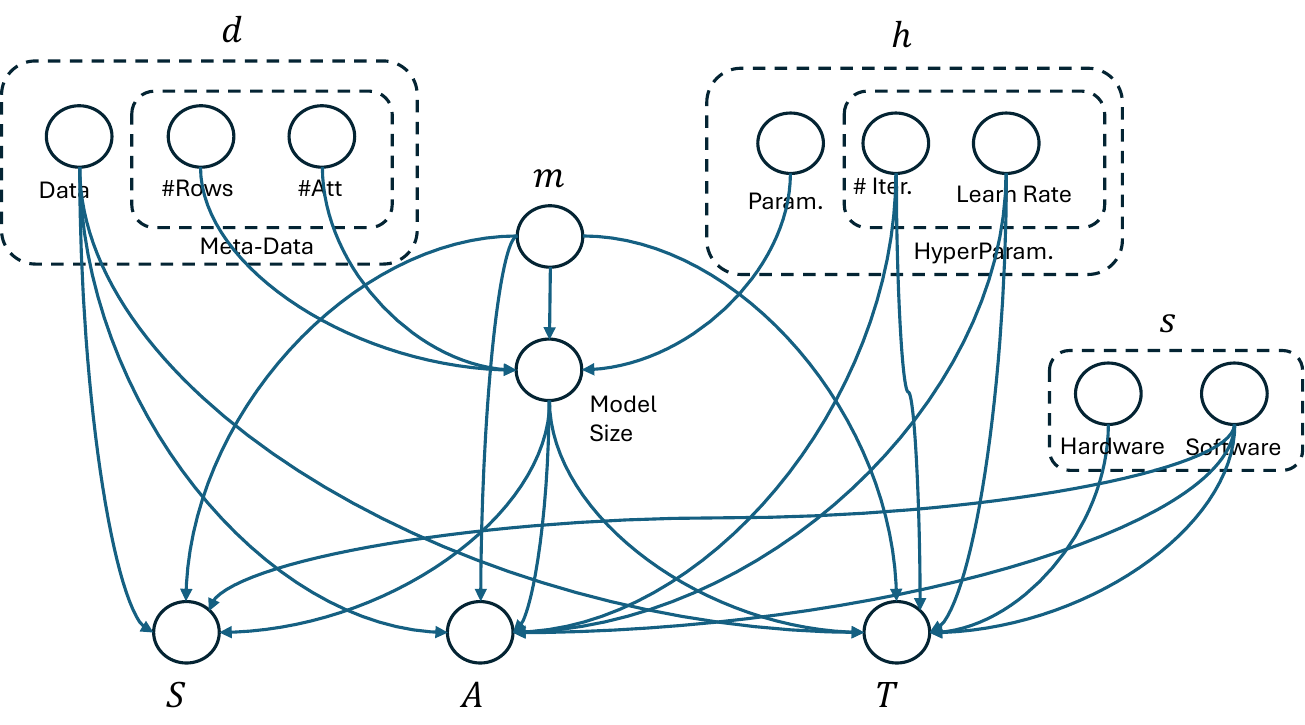}
   \caption{Outline of the causal graph enabling the causally-informed exploration and analysis of a  benchmark}
   \label{fig:cb_causal}
   \vspace{-0.1in}
\end{figure}

\paragraph{Causally-Informed Exploration and Analysis of Benchmark Runs} A user can visualize and explore a benchmark run, consisting of multiple benchmark scenarios, instrumented and executed on the same hardware by the same user. This involves slicing and dicing a benchmark run based on the datasets and models and comparing the different metric results and resource consumption. The entire benchmark run or its various subsets can be downloaded by the user for external analysis and visualization (Figure \ref{fig:cb_causal}). In addition, the user can create {\em virtual} benchmark runs by declaring a new benchmark context and collecting all compatible benchmark scenarios that have been instrumented, executed, and recorded in CausalBench at different times, potentially by different users. This enables the user to explore the performance of the models on different hardware/software settings.

Since accuracy, timing, and resource usage of the models may be impacted by the properties of the data, underlying parameter/hyper-parameter settings, as well as hardware/software configurations, CausalBench provides services to (a) disaggregate, de-bias, and explain the various factors impacting accuracy, time, and/or resource performance of the benchmark runs, as well as (b) propose new scenarios to execute to obtain a more robust understanding of the model performance. 

Figure~\ref{fig:cb_causal} provides the outline of the causal graph that forms the basis of these causally-informed exploration and analysis services. More specifically, CausalBench leverages a priori causal knowledge, described in the form of a causal graph, to boost the representational ability  and achieve better explanations and recommendations. Specifically, given a causal graph (possibly enriched by data-driven causal impact analysis~\cite{cauEffEst,cauRec, Sheth2023CausalDisentanglement, Sheth2022CausalDisentanglementSISAP, Sheth2024CausalityGuided, Li2023CTT})  describing the underlying causal relationships among the various factors impacting performance, CausalBench integrates this information into the learning process to ensure that explanations and recommendations are causally-robust. The causally-informed exploration and analysis services provided by CausalBench includes the following:
\begin{itemize}[leftmargin=*]
    \item {\em Causal impact and sensitivity analysis:} The benchmark data are analyzed through a causal effect discovery algorithm~\cite{caualgs1, caualgs2} to quantify the impacts of various factors on the target accuracy, time, or resource usage in a given context. 
    \item {\em Causal ranking and exploration:} Given a set of potentially conflicting decision parameters, the causal graph is also used to identify a non-dominating (pareto-optimal) subset of the runs that best highlight/explain the underlying trade-offs.
    \item {\em Causal prediction (with knowledge transfer):} Given a causal model and a benchmark of runs, CausalBench provides causally-informed performance predictions under new settings~\cite{wei2023transfer, Choi2022ReviewBased}. CausalBench tackles data sparsity through causally-informed knowledge transfer across simulation contexts, by disaggregating shareable and non-shareable information relying on the underlying causal structure.
    \item {\em Causal recommendations:} CausalBench aggregates the above impact analysis, ranking, and prediction services into a causally-informed recommendation service, which recommends additional benchmark configurations to execute.  
\end{itemize}

\section{Demonstration Scenarios \label{sec:demo}}
The demonstration scenarios include (a) dataset, model, and metric registration, (b) exploration, (c) benchmark context declaration, (d) benchmark instrumentation and execution, and (e) benchmark result exploration. Three sample scenarios are outlined next: 

\begin{itemize}[leftmargin=*]
     \item Scenario 1: User registers → logs in → retrieves the API key → downloads CB → implements their own dataset/model/metric → uploads the items, creating a submission → runs the context → posts the results → makes the results public and obtains DOI.
     \item Scenario 2: User logs in → browses through an array of datasets, metrics, models, and contexts by sorting and filtering → creates a benchmark context by selecting several sets, models, and metrics → downloads and instruments the benchmark → executes the benchmark → uploads results to CB and obtains DOI.
     \item Scenario 3: User logs in → creates a virtual benchmark context by selecting several datasets and models → CB aggregates and presents matching benchmark runs → user slices-and-dices the runs and obtains causal explanations and causally-informed recommendations for additional benchmark contexts to execute.
\end{itemize}
CausalBench is accessible at \cite{cbLink} and a 3-minute video recording showcasing the major features of CausalBench is available at \cite{vidLink}.

\section{Conclusions}
In this paper, we introduced CausalBench, a platform designed to support the benchmarking of causal learning models by facilitating scientific collaboration on novel algorithms, datasets, and metrics and promoting reproducibility in causal learning research. 

\bibliographystyle{ACM-Reference-Format}
\balance
\bibliography{bibcikm}

\end{document}